\documentclass[runningheads]{llncs}
\usepackage{graphicx}
\usepackage{comment}
\usepackage{amsmath,amssymb} % define this before the line numbering.
\usepackage{color}

\usepackage{subcaption}
\usepackage{booktabs}
\usepackage{array}
\usepackage{xspace}
\usepackage{cases}
\usepackage{multirow,boldline}
\usepackage{url}
\usepackage{xcolor}
\usepackage{paralist}
\usepackage[normalem]{ulem}
\usepackage{enumitem}
\usepackage{comment}
\usepackage[lined,ruled,linesnumbered]{algorithm2e}
\usepackage{algorithmic}
\usepackage{tabularx}
\usepackage[export]{adjustbox}

\newcommand{\etal}{\textit{et al.}}

\newcommand{\mycaption}[2]{\caption{\textbf{#1.}~#2}}
\usepackage{pifont}
\newcommand{\cmark}{\ding{51}}%
\newcommand{\xmark}{\ding{55}}%
\begin{document}
% \renewcommand\thelinenumber{\color[rgb]{0.2,0.5,0.8}\normalfont\sffamily\scriptsize\arabic{linenumber}\color[rgb]{0,0,0}}
% \renewcommand\makeLineNumber {\hss\thelinenumber\ \hspace{6mm} \rlap{\hskip\textwidth\ \hspace{6.5mm}\thelinenumber}}
% \linenumbers
\pagestyle{headings}
\mainmatter
\def\ECCVSubNumber{3376}  % Insert your submission number here

\title{Orientation-aware Vehicle Re-identification with Semantics-guided Part Attention Network} % Replace with your title

% INITIAL SUBMISSION 
\begin{comment}
\titlerunning{ECCV-20 submission ID \ECCVSubNumber} 
\authorrunning{ECCV-20 submission ID \ECCVSubNumber} 
\author{Anonymous ECCV submission}
\institute{Paper ID \ECCVSubNumber}
\end{comment}
%******************

% CAMERA READY SUBMISSION
%\begin{comment}
\titlerunning{Orientation-aware Vehicle Re-ID with Semantics-guided Part Attention Net}
% If the paper title is too long for the running head, you can set
% an abbreviated paper title here
%
\begin{comment}
\author{First Author\inst{1}\orcidID{0000-1111-2222-3333} \and
Second Author\inst{2,3}\orcidID{1111-2222-3333-4444} \and
Third Author\inst{3}\orcidID{2222--3333-4444-5555}}

%
\authorrunning{F. Author et al.}
% First names are abbreviated in the running head.
% If there are more than two authors, 'et al.' is used.
%
\institute{Princeton University, Princeton NJ 08544, USA \and
Springer Heidelberg, Tiergartenstr. 17, 69121 Heidelberg, Germany
\email{lncs@springer.com}\\
\url{http://www.springer.com/gp/computer-science/lncs} \and
ABC Institute, Rupert-Karls-University Heidelberg, Heidelberg, Germany\\
\email{\{abc,lncs\}@uni-heidelberg.de}}
\end{comment}

\author{
Tsai-Shien Chen\inst{1,2} \and
Chih-Ting Liu\inst{1,2} \and
Chih-Wei Wu\inst{1,2} \and
Shao-Yi Chien\inst{1,2}}

\institute{Graduate Institute of Electronic Engineering, National Taiwan University \\ \and
NTU IoX Center, National Taiwan University \\
\email{\{tschen, jackieliu, cwwu\}@media.ee.ntu.edu.tw},  
\email{sychien@ntu.edu.tw}}

%******************
\maketitle
\begin{abstract}
Vehicle re-identification (re-ID) focuses on matching images of the same vehicle across different cameras. 
It is fundamentally challenging because differences between vehicles are sometimes subtle.
While several studies incorporate spatial-attention mechanisms to help vehicle re-ID, they often require expensive keypoint labels or suffer from noisy attention mask if not trained with expensive labels.
In this work, we propose a dedicated Semantics-guided Part Attention Network (SPAN) to robustly predict part attention masks for different views of vehicles given only image-level semantic labels during training. With the help of part attention masks, we can extract discriminative features in each part separately. Then we introduce Co-occurrence Part-attentive Distance Metric (CPDM) which places greater emphasis on co-occurrence vehicle parts when evaluating the feature distance of two images. Extensive experiments validate the effectiveness of the proposed method and show that our framework outperforms the state-of-the-art approaches.
\end{abstract}

\section{Introduction}

\begin{figure}[t]
	\centering
    \includegraphics[width=0.85\textwidth]{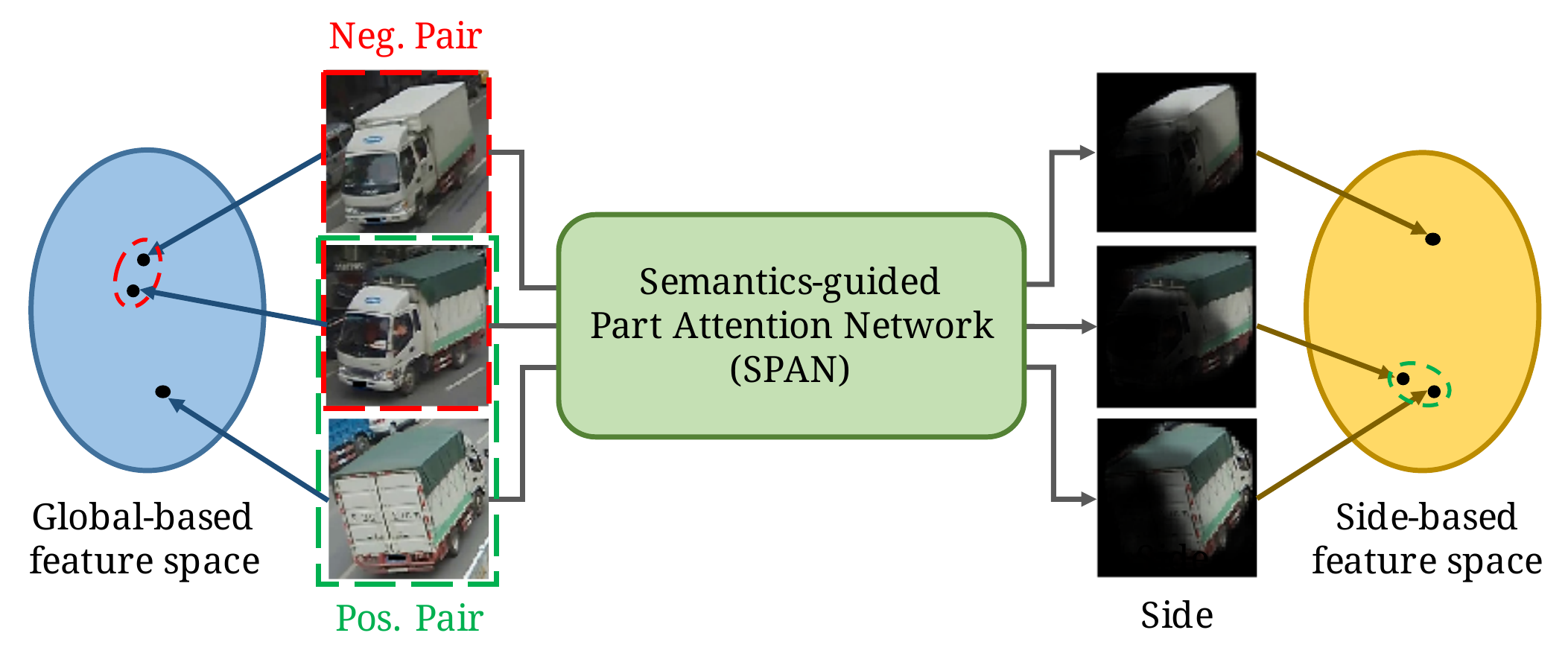}
    \mycaption{Concept illustration of Semantics-guided Part Attention Network}{
    The example images show intra-class difference and inter-class similarity in the vehicle re-ID problem. It is challenging to separate the negative images merely based on global feature due to the similar car model and viewpoint. In this example, it is easier to distinguish two vehicles by the side-based feature. This motivates us to generate the part (view) attention maps and then emphasize the feature of the co-occurrence vehicle parts for better re-ID matching. }
    \label{fig:intro}
\end{figure}

Vehicle re-identification (re-ID) aims to match vehicle images in a camera network. 
Recently, this task has drawn increasing attention due to practical applications such as urban surveillance and traffic flow analysis.
While deep Convolutional Neural Networks (CNN) have shown remarkable performance in vehicle re-ID over the years~\cite{VeRi-776-1,VeRi-776-2,cityflow}, various challenges still hinder the performance of vehicle re-ID.
One of them is that a vehicle captured from different viewpoints usually has dramatically different visual appearances. 
On the other hand, two different vehicles of the same color and car model are likely to have very similar appearances.
As illustrated in the left part of Fig.~\ref{fig:intro}, it is challenging to distinguish vehicles by comparing the features extracted from the whole vehicle images.
In such case, the minor differences in specific parts of vehicle such as decorations or license plates would be a great benefit to identifying two vehicles.
Furthermore, when two vehicles are presented in different orientations, a desired vehicle re-ID algorithm should be able to focus on the parts (views) that both appear in the two vehicle images.
For example, in the right part of Fig.~\ref{fig:intro}, it is easier to distinguish the vehicles by comparing their side views. To reach this idea, we divide it into two steps.

The first step is to extract the feature from specific parts of vehicle images. A number of work has been proposed to achieve this purpose by learning orientation-aware features.
Nonetheless, existing methods either rely on expensive vehicle keypoints as guidance to learn an attention mechanism for each part of a vehicle~\cite{orientation,advkeypt} or use only viewpoint labels but produce noisy and unsteady attention outcome which will thus hinder the network to learn subtle differences between vehicles~\cite{VAMI}.
In this paper, we introduce the \textit{Semantics-guided Part Attention Network (SPAN)} to generate attention masks for different parts (front, side and rear views) of a vehicle. 
As shown in Fig.~\ref{fig:intro}, our SPAN learns to produce meaningful attention masks.
The masks not only help disentangle features of different viewpoints but also improve the interpretability of our learning framework. 
It is also worth noting that, instead of expensive keypoints or pixel-level labels for training, our SPAN requires only \textit{image-level viewpoint labels} which are much easier to be derived from known camera pose and traffic direction.

For the second step, we design a \textit{Co-occurrence Part-attentive Distance Metric (CPDM)} to better utilize the part features when measuring the distance of images.
The intuition of this metric is that the network should focus on the parts (views) that both appear in the compared vehicle images.
Therefore, the proposed metric allows us to automatically adjust the importance of each part feature distance according to the part visibility in two compared vehicle images.

We conduct experiments on two large-scale vehicle re-ID benchmarks and demonstrate that our method outperforms current state-of-the-arts.
Ablation studies prove that the attention masks generated by SPAN extract helpful part features and our CPDM can better utilize the global and part features to improve the re-ID performance.
Moreover, qualitative results show that our SPAN can robustly generate meaningful attention maps on vehicles of different types, colors, and orientations.
We now highlight our contributions:
(1) We propose a Semantics-guided Part Attention Network (SPAN) to generate robust part attention masks which can be used to extract more discriminative features.
(2) Our SPAN only needs image-level viewpoint labels instead of expensive keypoints or pixel-level annotations for training.
(3) We introduce the Co-occurrence Part-attentive Distance Metric (CPDM) to facilitate vehicle re-ID by focusing on the parts that jointly appear in the compared images.
(4) Extensive experiments on public datasets validate the effectiveness of each component and demonstrate that our method performs favorably against state-of-the-art approaches.

\begin{comment}
\begin{enumerate}
\item We propose a Semantics-guided Part Attention Network (SPAN) to generate robust part attention masks which can be used to extract more discriminative features for vehicle re-ID.
\item Our SPAN only needs image-level viewpoint labels instead of expensive keypoints or pixel-level annotations for training.
\item We introduce the Co-occurrence Part-attentive Distance Metric (CPDM) to facilitate vehicle re-ID by focusing on the parts that jointly appear in the compared vehicle images.
\item Extensive experiments on public datasets validate the effectiveness of each component and demonstrate that our method performs favorably against state-of-the-art approaches.
\end{enumerate}
\end{comment}
\section{Related Work}

\paragraph{\textbf{Re-Identification (re-ID).}}
Re-identification studies the problem of identifying identities in different camera views.
There are large numbers of studies that focus on re-identifying human~\cite{scalable,part-aligned,stripe,context-aware} and vehicles~\cite{pathLSTM,orientation,VAMI,RAM}.
Most re-ID methods can be categorized into two types: feature learning and distance metric learning.
Feature learning methods~\cite{part-aligned,context-aware,region-guide,stripe,semantic,scalable} aim to learn a more discriminative embedding space.
Distance metric learning methods~\cite{crossview,onlinelocal,quadruplet,largescale,oneshot} design distance functions for comparing features of two images.
In this work, we design an orientation-aware feature extraction network as well as an orientation-aware distance metric for solving the vehicle re-ID problem.

\paragraph{\textbf{Vehicle Re-Identification.}}
Vehicle re-ID has received more attention for the past few years due to the releases of large-scale annotated vehicle re-ID datasets. 
Liu~\etal~\cite{VeRi-776-1,VeRi-776-2} released a high-quality multi-viewed VeRi-776 dataset. Tang~\etal~\cite{cityflow} proposed a city-scale traffic camera CityFlow dataset. With several datasets, numerous vehicle re-ID methods have been proposed recently.
Some methods use CNN model to tackle the vehicle re-ID problem~\cite{pathLSTM,RAM,cityflow}. However, those methods lack spatial guidance and could be hard to distinguish two similar vehicles with only subtle difference.
In contrast, the others adopt the extra information, such as viewpoint or keypoint labels, to generate spatial attentive features.
Wang~\etal~\cite{orientation} and Khorramshahi~\etal~\cite{advkeypt} used 20 vehicle keypoints to generate attention maps by categorizing keypoints into four groups which respectively represent front, rear, left or right view of vehicle. Yet, the keypoint information is hard to acquire in real-world scenarios. Also, the keypoint is insufficient to cover all crucial features.
Zhou~\etal~\cite{VAMI} proposed a viewpoint-aware attention model to produce attention map for different viewpoints and further generate multi-view features from single view input image. However, due to the lack of direct supervision on the generated attention maps, the attention outcomes are noisy and would unfavorably affect the learning of network. 
In contrast, we design a dedicated network and adopt specific loss functions to supervise the generation of attention maps. Moreover, our network only requires image-level viewpoint labels rather than keypoint labels during training.

\paragraph{\textbf{Visibility-Aware Features.}}
Utilizing visibility-aware features has gained growing interest considering that there are lots of occluded images in real-world scenarios. Sun~\etal~\cite{where} and Miao~\etal~\cite{occlude} pre-define several regions among whole images by horizontally or vertically partitioning the images and then produce the confidence score for each region to represent their visibility. However, the visibility of pre-defined region is hard to represent its importance for re-ID matching. For example, the highly visible regions but containing mostly background would be overemphasized while the smaller regions but containing some critical appearances would be neglected. To avoid the issue mentioned above, in this work, we directly use the visibility of specific parts of vehicle to represent its importance. Note that it is only possible when the specific parts are accurately located.
\begin{figure}[t!]
	\centering
    \includegraphics[width=\textwidth]{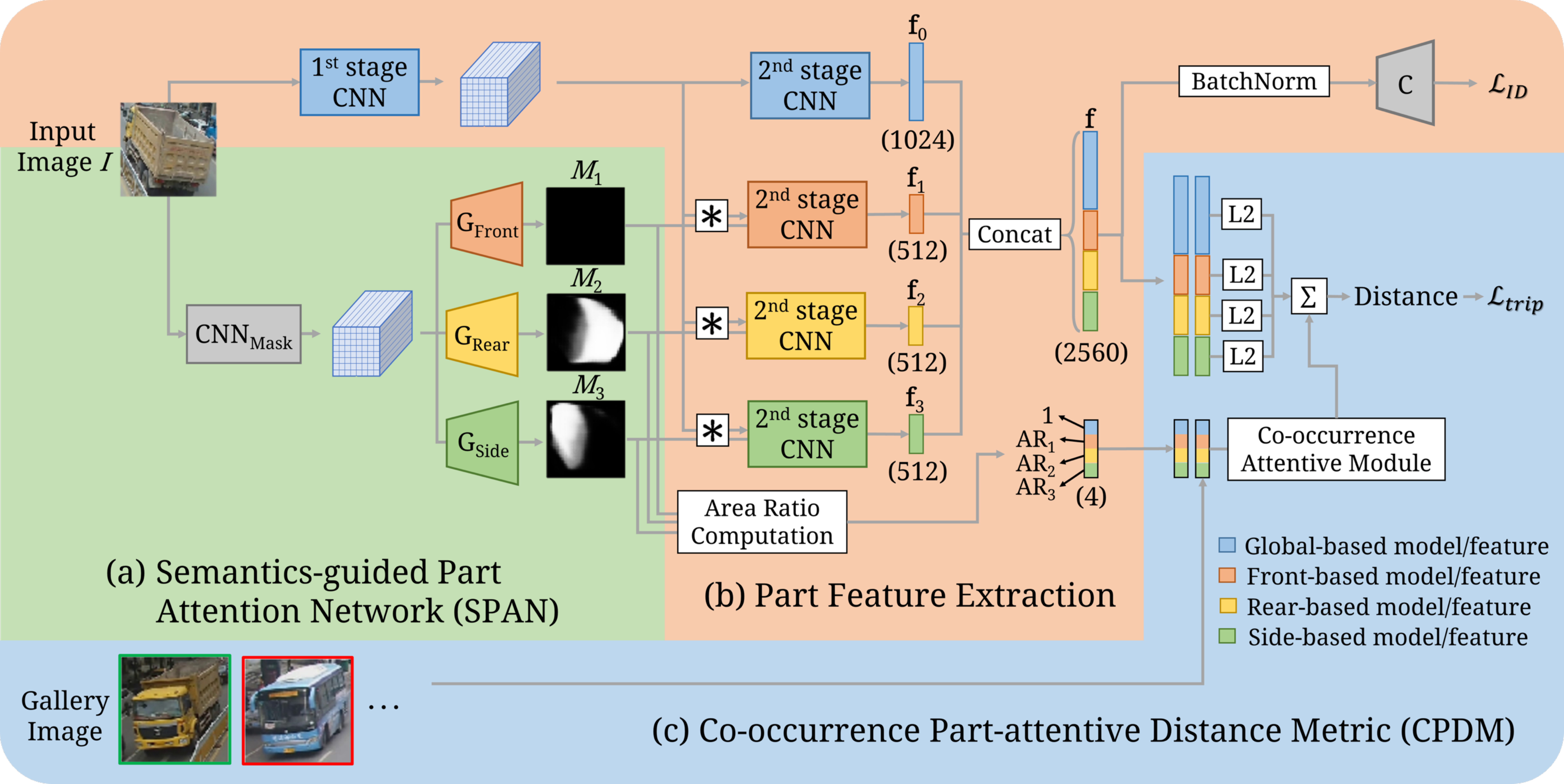}
    \mycaption{Architecture of our proposed framework}{
    (a) Semantics-guided Part Attention Network (SPAN) generates the attention masks for each part (view) of a vehicle image. (b) With the attention masks generated by SPAN, Part Feature Extraction produces one global and three part attentive features which are then concatenated into a representative feature. (c) Co-occurrence Part-attentive Distance Metric (CPDM) calculates a weighted feature distance with emphasis on the vehicle parts that appear in both compared images.}
    \label{fig:overview}
\end{figure}

\section{Proposed Method}
The proposed learning framework for vehicle re-ID consists of three sub-modules as depicted in Fig.~\ref{fig:overview}.
First, we learn a Semantics-guided Part Attention Network (SPAN) to predict the attention masks for each part (view) of a vehicle in Sec.~\ref{OrienSeg}.
Then, in Sec.~\ref{FeatExt}, we apply the attention masks to our main feature extraction network to generate part features in addition to the global features.
During both training and inference, the global and part features are combined to evaluate feature distance between two vehicle images with our proposed Co-occurrence Part-attentive Distance Metrics (CPDM) in Sec.~\ref{DistMet}.
Last, the overall model learning scheme of our framework is introduced in Sec.~\ref{Train}.

\subsection{Semantics-guided Part Attention Network}
\label{OrienSeg}
The goal of our Semantics-guided Part Attention Network (SPAN) is to generate a set of attention masks for different parts (e.g. front, side, and rear view) of a vehicle image. 
An intuitive approach would be to train a segmentation network with pixel-wise view labels to predict segmented part masks.
However, pixel-level annotation is expensive to obtain in real-world data.
Instead, we turn to the image-level semantic labels, such as the viewpoint of vehicles which are much easier to be derived from known camera pose and the traffic direction, to learn our attention network.
Given a vehicle image $I$, we define its corresponding semantic label vector as $\boldsymbol{l} \in \mathcal{R}^3$.
The semantic label $\boldsymbol{l}$ is encoded from its viewpoint.
Its elements represent whether the front, rear or side view of image $I$ are visible or not, respectively.
To be more specific, $l_i = 1$ if the $i^{th}$ view is visible, while $l_i = 0$ if it is not. 
For example, for a vehicle image with the front-side viewpoint, its semantic label vector $\boldsymbol{l}$ will be assigned with $[1, 0, 1]$.

As shown in Fig.~\ref{fig:overview}~(a), our network predicts the attention masks of front, rear and side views $M_1, M_2, M_3$ with a shared feature extractor ${CNN}_{Mask}$ and three mask generators $G_{Front}$, $G_{Rear}$, and $G_{Side}$. 
To ensure our SPAN generating ideal masks, we meticulously design a novel loss function, named mask reconstruction loss, with two auxiliary losses to supervise the learning of network.

\begin{figure}[t]
	\centering
    \includegraphics[width=0.75\textwidth]{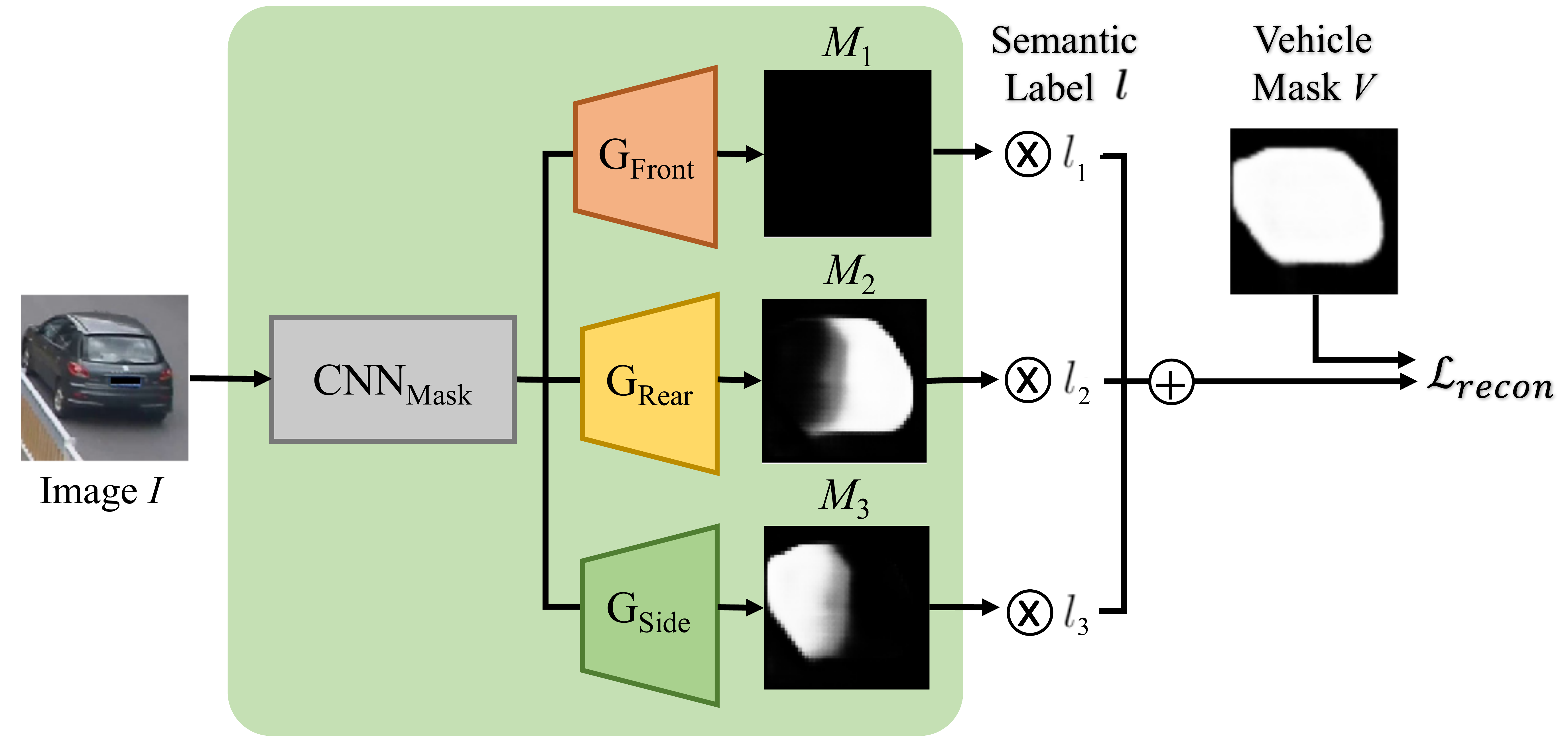}
    \mycaption{Mask Reconstruction Loss}{
    The part masks selected by semantic label should jointly reconstruct the whole foreground vehicle mask.}
    \label{fig:recon}
\end{figure}

\paragraph{\textbf{Mask Reconstruction Loss.}}
As illustrated in Fig.~\ref{fig:recon}, the main idea of mask reconstruction loss is that the attention masks selected by corresponding semantic labels should jointly reconstruct the foreground mask of a vehicle.
For instance, if the image is with rear-side viewpoint, the rear and side masks should jointly reconstruct the whole vehicle foreground mask to the greatest extent possible.
To this end, we first need the foreground mask of each vehicle image, which is also automatically generated by our deep segmentation network trained with the preliminary results by GrabCut~\cite{grabcut} as the target. The detail of generating foreground masks is shown in the supplementary material; notes that any manually annotated pixel-level label is \textbf{not} required here. Thus, with the foreground masks (denoted as $V$), our mask reconstruction loss can be written as:
\begin{equation}
\label{eq:loss_recon}
    \mathcal{L}_{recon} = \|V - \sum_{i=1}^{3}{(l_i\times M_i)}\|_2,
\end{equation}
which represents the mean square error (MSE) between the foreground mask and the generated mask gated by the semantic label.

\paragraph{\textbf{Area Constraint Loss.}}
While imposing the mask reconstruction loss, we note that the training is unstable and often leads to undesired results.
Take the qualitative result in Fig.~\ref{fig:ablation_loss} ``w/o~$\mathcal{L}_{area}$'' as example, we observe that, for a vehicle image with two visible views, the network only uses single representative mask generator to predict the whole vehicle mask.
To prevent network from cheating, we design the area constraint loss to limit the maximum area of each predicted attention mask.
Here, we define the area of mask as its L1-norm (sum of all elements) and also define the maximum area ratio of $i^{th}$ view for a semantic label $\boldsymbol{l}$ as $a_{\boldsymbol{l},i}$. 
Our area constraint loss can be formulated as:
\begin{equation}
\begin{aligned}
\label{eq:loss_area}
 \mathcal{L}_{area} = \sum_{i=1}^{3}{\left[\dfrac{\|M_i\|_1}{\|V\|_1}-a_{\boldsymbol{l},i} \right]_+},
\end{aligned}
\end{equation}
where $\|\cdot\|_1$ represents L1-norm of a given mask. 
$[\cdot]_+$ is the hinge function since we only penalize the mask with the area ratio (over the whole foreground mask) larger than our expected ratio.
For the setting of max area ratio $a$, the ratio of invisible parts should be 0 intuitively while the ratio of visible part should be 1 for images with merely one visible views. For images with two visible views, the ratio of each view should be set within the range from 0.5 to 1.% and should be within the range from 0.5 to 1 for images with two visible views.

\paragraph{\textbf{Spatial Diversity Loss.}}
In addition to the situation mentioned above, we observe other unfavorable results. Such as the qualitative result in Fig.~\ref{fig:ablation_loss} ``w/o~$\mathcal{L}_{div}$'', for a vehicle image with two visible views, the two corresponding mask generators may predict whole vehicle masks with values of $0.5$.
Therefore, similar to Li~\etal~\cite{diversity}, we introduce a spatial diversity loss to restrict the overlapped area between masks of different views with the following formulation:
\begin{equation}
\label{eq:loss_div}
 \mathcal{L}_{div} = \sum_{(i,j)\in P}{[(M_{i}\cdot M_{j}) - m_{i, j}]_+},
\end{equation}
where $m_{i,j}$ is the margin representing the tolerable overlapped area between $i^{th}$ and $j^{th}$ view and $P$ is the set of all view index pairs.
For two mutually exclusive views, such as front and rear, the margin is set to 0 intuitively. For two adjacent views, such as front and side, the margin parameter is set to a positive value to tolerate the overlapped situation (e.g. front-side view mirror and headlight could be hard to uniquely assign to either front or side view).

\begin{figure}[t]
	\centering
    \includegraphics[width=0.6\textwidth]{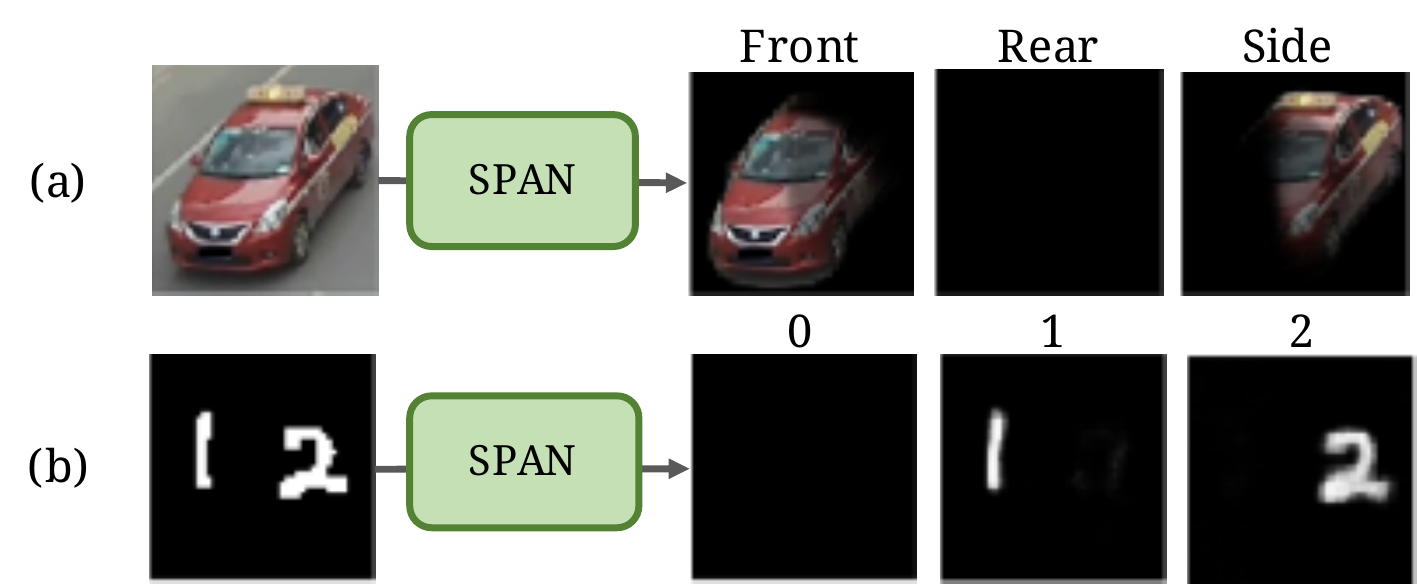}
    \mycaption{Illustration of general-purposed SPAN}{
    (a)(b) show the output masks of the vehicle and multi-digit images with different semantic labels respectively. Our proposed SPAN is able to learn to generate the part attention map or localization only given the image-level semantic labels.}
    \label{fig:general_SPAN}
\end{figure}

\paragraph{\textbf{Discussion.}}
SPAN is general-purposed and can be extended to weakly-supervised segmentation which has much weaker supervision setting than regular segmentation because it only requires image-level label for training. It can also well perform on other datasets besides to vehicle images. Fig.~\ref{fig:general_SPAN} shows the example results on the multi-digit dataset based on MNIST~\cite{MNIST} created by ourselves. For the multi-digit dataset, the semantic label represents which digit is visible in the image. Hence, the network can learn to generate the localization of each digit.
%The proposed SPAN is general-purposed and can also well perform on other datasets. Besides on vehicle image dataset, Fig.~\ref{fig:general_SPAN} shows the example results on the multi-digit dataset based on MNIST~\cite{MNIST} created by ourselves. For the multi-digit dataset, the semantic label represents which digit is visible in the image. Hence, the network can learn to generate the localization of each digit.

\subsection{Part Feature Extraction}
\label{FeatExt}
With the attention masks generated by our SPAN, we design a part feature extraction module to learn orientation-aware features for vehicle re-ID. 
As shown in Fig.~\ref{fig:overview} (b), the module includes two convolution stages. 
The $1^{st}$-stage CNN transforms input images into $1^{st}$-stage feature maps. 
Then, four distinct $2^{nd}$-stage CNNs respectively dedicated for extracting global-based, front-based, rear-based and side-based features follow the previous stage. 
The global-based model simply takes $1^{st}$-stage feature map as input and generates the global feature $\textbf{f}_0$. 
The other three branches apply the part attention masks to the $1^{st}$-stage feature map by element-wise matrix multiplication and then extract part features $\textbf{f}_1$, $\textbf{f}_2$ and $\textbf{f}_3$ by corresponding $2^{nd}$-stage CNNs.
With one global feature and three part features, unlike previous methods~\cite{orientation,VAMI,advkeypt} which embed all part features into one unified vector by additional network, our network simply concatenates them into one representative feature $\textbf{f}$ to best utilize all possible features for vehicles.

\begin{figure}[t]
	\centering
    \includegraphics[width=0.75\textwidth]{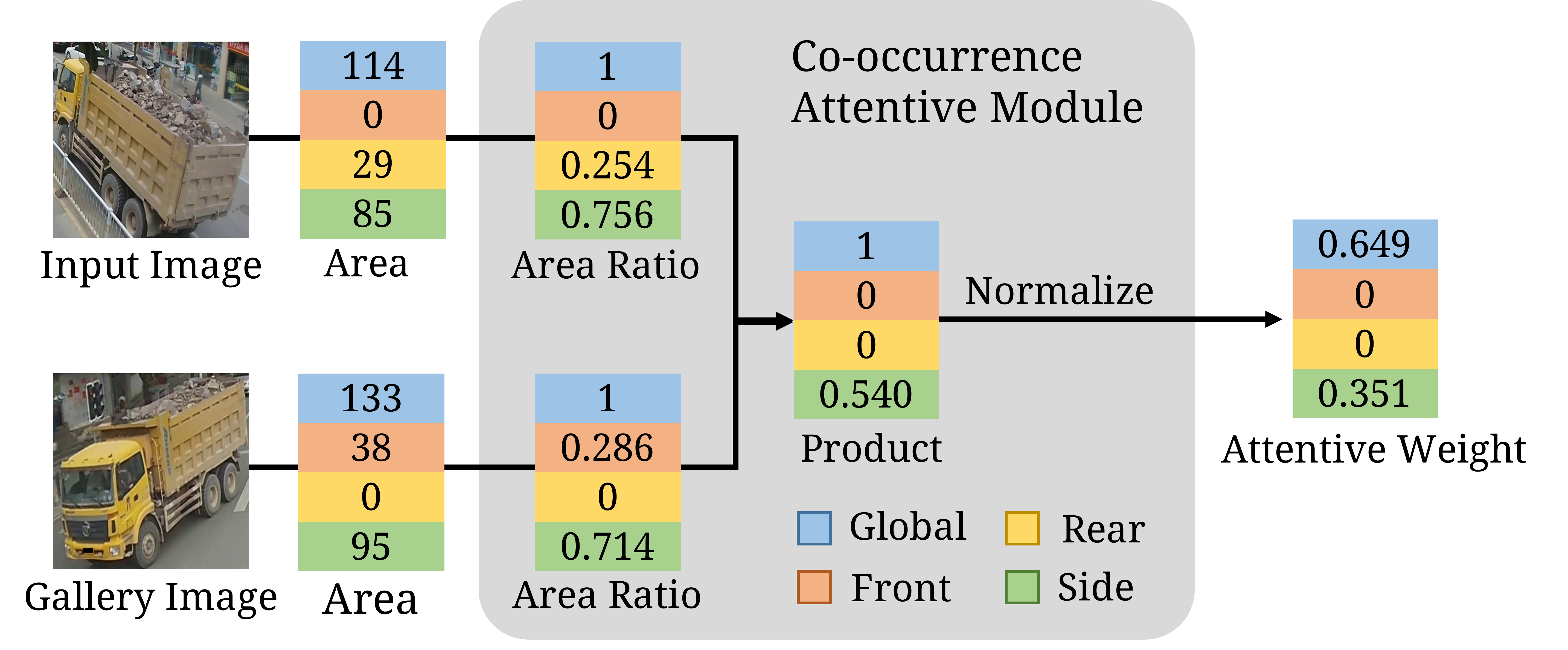}
    \mycaption{Co-occurrence Attentive Module}{
    To correctly recognize these two positive images, the feature of side view (co-occurrence view) should be emphasized, while front and rear feature should be relatively neglected. Co-occurrence attentive module is able to re-weigh the importance of each view accordingly.}
    \label{fig:distEx}
\end{figure}

\subsection{Co-occurrence Part-attentive Distance Metric}
\label{DistMet}
To fully utilize the part features extracted by our SPAN and part feature extraction module, we design the Co-occurrence Part-attentive Distance Metric (CPDM) for both training the CNN and matching images during inference.
We note that, in addition to the global feature, the features of the same visible parts on different vehicles are also critical for re-ID.
Moreover, the co-occurrence part with greater area ratio often represents higher clarity or is likely to include more key features in the original image.
Therefore, we develop the Co-occurrence Attentive Module to re-weigh the importance of different feature distances by comprehensively considering the area ratio of each view in both images.
Fig.~\ref{fig:distEx} illustrates an example of Co-occurrence Attentive Module.
Given a vehicle image, we first compute the area of global, front, rear and side view by calculating the L1-norm of the attention masks generated by SPAN (the area of global view is defined as the summation of the ones of front, rear and side view). 
The area ratios of each view are then normalized by the global area.
We denote the area ratio of $i^{th}$ view in image $I_a$ as $AR_{a,i}$. 
For arbitrary two images $I_a$ and $I_b$, the attentive weight of $i^{th}$ view $w_{(a,b),i}$ can be written as:
\begin{equation}
\label{eq:orien_att}
 w_{(a,b),i} = \frac{AR_{a,i}\times AR_{b,i}}{\sum_{i=0}^{3}{AR_{a,i}\times AR_{b,i}}}. 
\end{equation}

Finally, we use the attentive weights to adjust the weighting for combining feature distances of all global and part features.
The final distance $Dist_{(a,b)}$ between two vehicle images $I_a$ and $I_b$ is calculated by:
\begin{equation}
\label{eq:distance}
 Dist_{(a,b)} = \sum_{i=0}^{3}{w_{(a,b),i}\times \|\textbf{f}_{a,i}-\textbf{f}_{b,i}\|_2},
\end{equation}
which is the weighted summation of feature euclidean distances in each view. 
\paragraph{\textbf{Discussion.}} For two images with completely disjoint views, the attentive weights are all 0 for front, rear, and side views. Hence, the distance between $I_a$ and $I_b$ will be fully determined by their global features $\textbf{f}_{a,0}$ and $\textbf{f}_{b,0}$.% which is same as baseline model (the model without CPDM).

\subsection{Model Learning Scheme}
\label{Train}
The learning scheme for our feature learning framework consists of two steps.
In the first step (Fig.~\ref{fig:overview} (a)), we optimize our SPAN with the following loss:
\begin{equation}
    \mathcal{L}_{step1} = \lambda_{recon}\mathcal{L}_{recon} + \lambda_{area}\mathcal{L}_{area} + \lambda_{div}\mathcal{L}_{div}.
\end{equation}
Instead of training SPAN end-to-end with the re-ID feature extractor network~\cite{end2end}, we train this network in advance because SPAN relies on clean viewpoint labels, which is not the case of our experimenting datasets.
As a result, we train SPAN with a smaller dataset than the original one but with cleaner viewpoint labels.

In the second stage, we optimize the rest of our network (Fig.~\ref{fig:overview} (b)(c)) with two common re-ID losses while SPAN is fixed. 
The first one for metric learning is the triplet loss ($\mathcal{L}_{trip}$)~\cite{wtriplet}, which is calculated based on the weighted distance introduced in Sec.~\ref{DistMet}.
The other loss for the discriminative learning is the identity classification loss ($\mathcal{L}_{ID}$)~\cite{discriminative}. 
The overall loss is computed as follows:
\begin{equation}
    \mathcal{L}_{step2} = \lambda_{trip}\mathcal{L}_{trip} + \lambda_{ID}\mathcal{L}_{ID}.
\end{equation}
During inference, given a query and a gallery image, we extract their features separately by SPAN and the part feature extraction module.
The distance of the query and gallery images are then computed by our CPDM for re-ID matching.
\section{Experiments}

\subsection{Datasets and Evaluation Metrics}
\label{datasets}
Our framework is evaluated on two benchmarks, VeRi-776~\cite{VeRi-776-1,VeRi-776-2} and CityFlow-ReID~\cite{cityflow}, which are two large-scale vehicle re-ID datasets with multiple viewpoints.
VeRi-776 dataset contains 776 different vehicles captured, which is split into 576 vehicles with 37,778 images for training and 200 vehicles with 11,579 images for testing.
Wang~\etal~\cite{orientation} released the annotated keypoints and viewpoint information for VeRi-776 dataset, which has been widely adopted by other work. 
In this paper, we only use the viewpoint labels to train our proposed SPAN. 
CityFlow-ReID is a subset of images sampled from the CityFlow dataset~\cite{cityflow}. 
It consists of 36,935 images of 333 identities in the training set and 18,290 images of another 333 identities in the testing set. 
However, the viewpoint information of CityFlow-ReID is not available. Thus, we utilize the SPAN pre-trained on VeRi-776 to generate corresponding attention masks.
Note that, though VehicleID~\cite{vehicleID} dataset is also a widely adopted benchmark, it only covers the images with front or rear viewpoint and cannot validate the effectiveness of our method. Hence, we would not use VehicleID   in the following experiments.

As in previous vehicle re-ID works, we employ the standard metrics, namely the cumulative matching curve (CMC) and the mean average precision (mAP)~\cite{scalable} to evaluate the results. 
We report the rank-1 accuracy (R-1) in CMC and the mAP for the testing set in both datasets.

\subsection{Implementation Details}
\label{details}
For our SPAN (Fig.~\ref{fig:overview} (a)), we adopt the former four blocks in ResNet-34~\cite{resnet} (\textit{conv1} to \textit{conv4}) as the feature extractor ($CNN_{mask}$) to extract the mid-level features which retain more spatial information than those after the last block (\textit{conv5}).
Afterwards, the feature map is fed into three generative blocks to generate the part masks. The detailed architecture of SPAN is shown in the supplementary material. 
This network is trained in advance on a subset of VeRi-776 dataset with balanced images in each viewpoint.
For optimizing SPAN with $\mathcal{L}_{step1}$, the coefficients $\lambda_{recon}$ and  $\lambda_{div}$ are set to $1$ and $\lambda_{area}$ is $0.5$.

For our part feature extraction (Fig.~\ref{fig:overview} (b)), we adopt ResNet-50~\cite{resnet} as our backbone which is split into two stages. The first four blocks ($conv1$ to $conv4$) are in the first stage and the last block ($conv5$) with one fully-connected layer are in the second stage to generate a 1024-d or 512-d feature vector. 
For optimizing with triplet loss ($\mathcal{L}_{trip}$), we adopt the $PK$ training strategy~\cite{metric}, where we sample $P=8$ different vehicles and $K=4$ images for each vehicle in a batch of size $32$. 
In addition, for training identity classification loss ($\mathcal{L}_{ID}$), we adopt a BatchNorm~\cite{bagoftricks} and a fully-connected layer as the classifier~\cite{bagoftricks,liu}. 
The training process lasts for 30,000 iterations with $\lambda_{trip}$ and $\lambda_{ID}$ all set to $1$ in $\mathcal{L}_{step2}$.

\subsection{Ablation Studies and Visualization}
\label{ablation}
In this section, to assess the effectiveness of our Semantics-guided Part Attention Network (SPAN) and Co-occurrence Part-attentive Distance Metric (CPDM), we conduct ablation studies quantitatively on VeRi-776 dataset and visualize the qualitative results of our attention masks compared with the existing methods.

\paragraph{\textbf{Loss Functions of Our SPAN.}}
\label{loss_study}
We adopt three loss functions to help generating steady and clear attention masks when training SPAN. To evaluate the influence of each loss function, we conduct experiments with multiple combinations of losses and report the re-ID results on VeRi-776 in Table~\ref{tab:ablation_loss} and the corresponding qualitative results of our part attention masks in Fig.~\ref{fig:ablation_loss}.

As listed in the first row in Table~\ref{tab:ablation_loss}, we show the baseline method which simply transferred the whole vehicle image into a $1024$-dim global feature and adopted euclidean distance as the feature distance metric. 
Except for the baseline method, all other methods in Table~\ref{tab:ablation_loss} adopt CPDM and utilize same architecture in SPAN but trained with different combinations of proposed loss functions. As shown in the second to fourth rows in Table~\ref{tab:ablation_loss} and the corresponding visualized attention masks in Fig.~\ref{fig:ablation_loss}, the re-ID performance of those methods are almost the same as the baseline owing to the unfavorable generated attention masks, which cannot benefit the part feature extraction and the following CPDM.
Only when simultaneously supervised by proposed three loss functions, our SPAN can generate clear and meaningful attention masks which can further improve the re-ID performance by a large margin as shown in the last row in Table.~\ref{tab:ablation_loss}.

\begin{center}
\begin{tabularx}{\textwidth}{*{2}{>{\centering\arraybackslash}X}}
   \centering
    \scalebox{0.85}{
    \begin{tabular}{l|ccc|cc}
        \hline
        \multirow{2}{*}{Method} & \multicolumn{3}{c|}{Training Loss} & \multicolumn{2}{c}{VeRi-776} \\ \cline{2-6}
        & $\mathcal{L}_{recon}$    & $\mathcal{L}_{area}$ & $\mathcal{L}_{div}$ & R-1 & mAP \\ \hline \hline
        Baseline                   & -      & -      & -      & 92.0 & 59.1 \\
        only $\mathcal{L}_{recon}$ & \cmark & \xmark & \xmark & 91.8 & 58.9 \\
        w/o $\mathcal{L}_{area}$   & \cmark & \xmark & \cmark & 92.1 & 59.2 \\ 
        w/o $\mathcal{L}_{div}$    & \cmark & \cmark & \xmark & 92.5 & 59.7 \\ \hline
    \textbf{SPAN(Ours)}            & \cmark & \cmark & \cmark & \textbf{93.9} & \textbf{68.6} \\
    \hline
    \end{tabular}} 
&
    \includegraphics[width=1.\linewidth,valign=m]{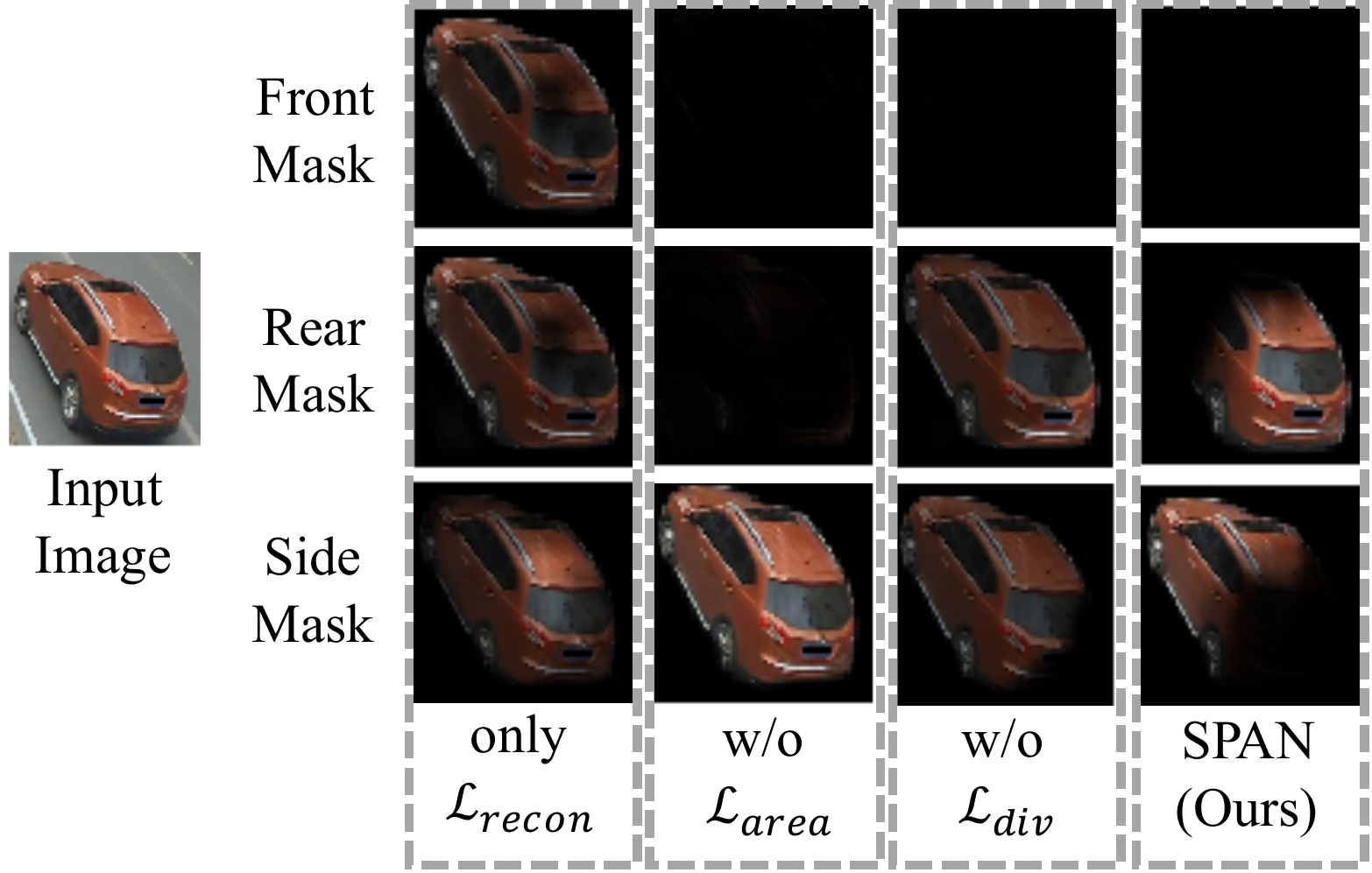} \\
    \captionof{table}{\textbf{Ablation study of the loss functions for training SPAN (\%).}}
    \label{tab:ablation_loss}
&     
    \captionof{figure}{\textbf{Results of SPAN training w/ different combinations of losses.}}
    \label{fig:ablation_loss}
\end{tabularx}
\end{center}

\paragraph{\textbf{Selection of Hyper-parameters in Loss Functions.}}
There are two hyper-parameters which should be selected for loss functions, including max area ratio $a$ in $\mathcal{L}_{area}$ and margin $m$ in $\mathcal{L}_{div}$. The physical meanings of selection have been discussed in Sec.~\ref{OrienSeg}. We finally choose $a = 0.7$ for the visible views of two-view images and $m = 0.04$ for two adjacent views based on the experimental results shown in the supplementary material.

\begin{figure*}[t!]
	\centering
    \includegraphics[width=1.\textwidth]{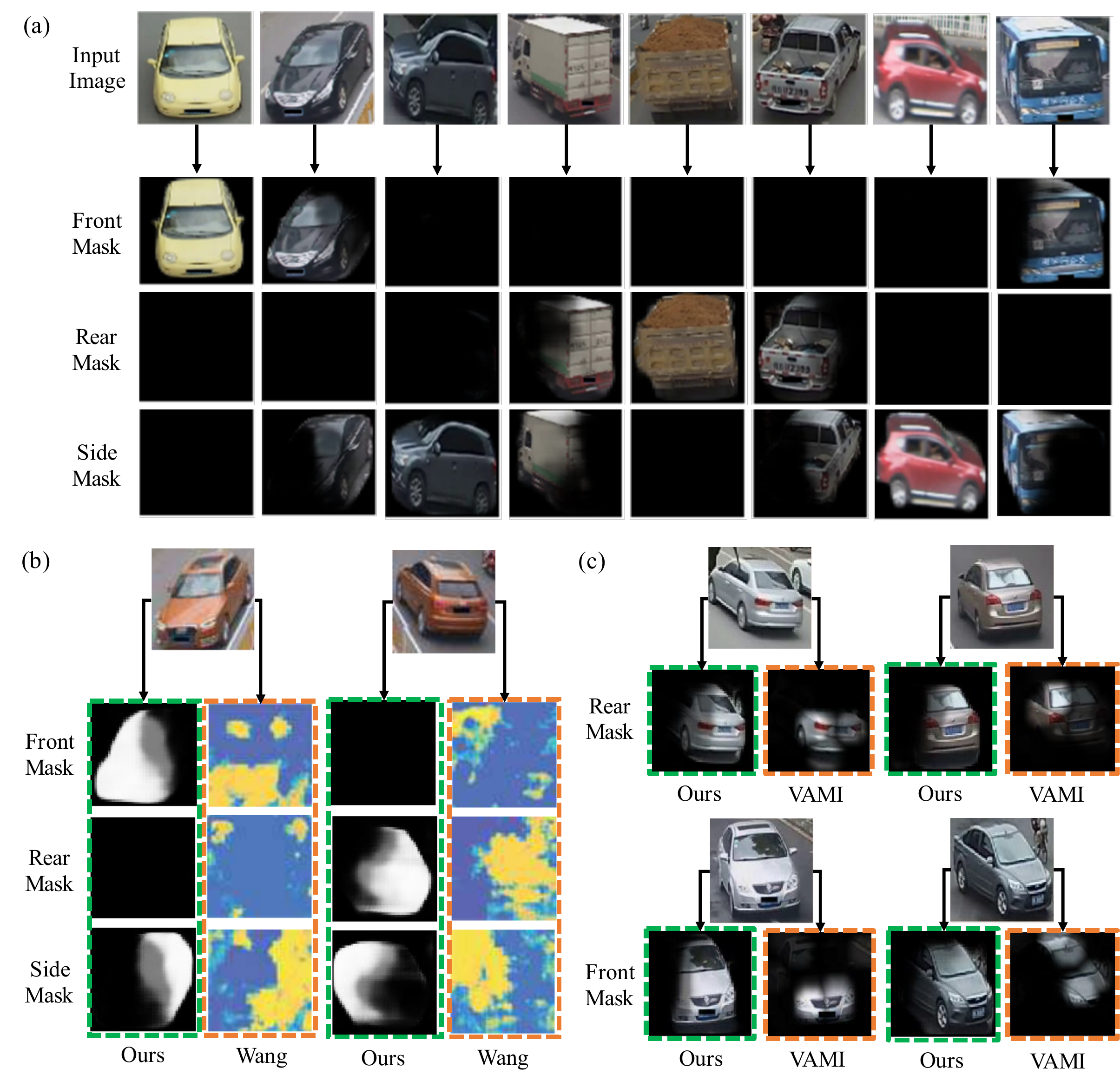}
    \mycaption{Qualitative Part Masks}{
    (a) shows some examples of the part masks generated by SPAN. Note that the demonstrated vehicles are all in different colors, types and orientations to verify the robustness of SPAN. (b)(c) show the comparison with Wang~\etal~\cite{orientation} and Zhou~\etal~\cite{VAMI} (VAMI) respectively. The attention maps generated by their methods are directly from their papers.}
    \label{fig:compare}
\end{figure*}

\paragraph{\textbf{Qualitative Results of Part Attention Masks.}}
\label{qualitative}
To verify the robustness of our proposed SPAN, we show some qualitative results of our part attention masks in Fig.~\ref{fig:compare} (a) and show the comparisons with Wang~\etal~\cite{orientation} and Zhou~\etal~\cite{VAMI} (VAMI) in Fig.~\ref{fig:compare} (b)(c), respectively.  In Fig.~\ref{fig:compare} (a), the produced masks from our SPAN can correctly cover all regional features which are belonging to their views while eliminates all redundant information such as features from background or other views. For example, headlights and front bumper are all covered in front mask, while door or background are not. In Fig.~\ref{fig:compare} (a), the demonstrated vehicles are all in different colors, types (sedan, SUV, pickup, truck, bus, etc.) and orientations, proving that SPAN is robust for various vehicles. 

In contrast, the attention masks generated by the previous work~\cite{orientation,VAMI} are more noisy and unsteady. As shown in the left half of Fig.~\ref{fig:compare} (b), the front mask generated by Wang~\etal~\cite{orientation} cannot cover the front windshield which possibly contains crucial features such as stickers or patterns. Also, in the right half of Fig.~\ref{fig:compare} (b), the front face of given vehicle image is not visible but the generated front attention mask fails to shield all features and instead activates on the background. 
The other example of generating unsteady masks is shown in Fig.~\ref{fig:compare} (c). Both rear and front masks generated by VAMI~\cite{VAMI} fail to consistently embed the rear or front windshield among different vehicle images, which will make the network hard to distinguish two images based on those part features.

\paragraph{\textbf{Component Analysis of the Proposed Model.}}
\label{component}
Here, we report the re-ID performances to evaluate the effectiveness of each sub-module in our proposed framework in Table~\ref{tab:ablation_proposed}. The first row demonstrates the baseline model which simply transfers the whole vehicle image into a global feature and uses standard euclidean distance to evaluate the distance between two vehicles. Next, based on our SPAN, we conduct two experiments with different aggregation techniques to combine the global and part features into one vector. The first one utilizes an additional fully-connected layer (FC) to embed the whole features, as shown in the second row in Table~\ref{tab:ablation_proposed} (SPAN w/ FC). The other directly concatenates all global and part features, as shown in the third row in Table~\ref{tab:ablation_proposed} (SPAN w/ Cat). It shows that compared to the baseline method, the performances are all boosted with the global and part features jointly be utilized. However, concatenating all the features can retain more part information, which achieves better performance for re-ID (from $59.1\%$ to $63.1\%$ than from $59.1\%$ to $60.3\%$ in mAP). Last, we report the results in the last row with the concatenated features and the usage of our CPDM, which is also our final proposed method (\textbf{SPAN w/ CPDM}). It shows that with our proposed method, the re-ID performance can outperform the baseline method by a large margin ($\mathbf{9.5}\%$ in mAP), proving that CPDM can better utilize the global and part features to measure the distance between two vehicles by enhancing the importance of the co-occurrence part.

\begin{table}[t]
    \centering
    \mycaption{Ablation studies of the proposed method in terms of R-1 and mAP ($\%$)}{The effectiveness analysis for each component including the usage of SPAN, feature aggregation methods (Agg.) and distance metrics (Dist.).}
    \label{tab:ablation_proposed}
    \begin{tabular}{l|ccc|cc}
    \hline
    \multirow{2}{*}{Method} & \multicolumn{3}{c|}{Sub-modules} & \multicolumn{2}{c}{VeRi-776} \\ \cline{2-6}
                        & SPAN   & Agg.    & Dist.& R-1  & mAP \\ \hline \hline   
    Baseline            & \xmark & -       & Euc. & 92.0 & 59.1   \\ 
    SPAN w/ FC          & \cmark & FC      & Euc. & 92.6 & 60.3   \\
    SPAN w/ Cat         & \cmark & Concat. & Euc. & 93.0 & 63.1   \\ \hline
    \textbf{SPAN w/ CPDM (Ours)} & \cmark & Concat. & CPDM & \textbf{94.0} & \textbf{68.9}   \\
    \hline
    \end{tabular}
\end{table}

\subsection{Comparison with the State-of-the-Arts}
\label{sota}
We compare our proposed framework with the state-of-the-art vehicle re-ID methods and report the results on VeRi-776 and CityFlow-ReID datasets in Table~\ref{tab:sota}.
Note that there are a few of recent works which cannot be fairly compared with ours due to different setting such as the usage of external vehicle re-ID dataset~\cite{joint}, manually annotated bounding boxes for crucial features~\cite{PRVR} and large-scale synthetic dataset with various kinds of pixel-level annotations~\cite{PAMTRI}. Therefore those works are not shown in our comparison in Table~\ref{tab:sota}.% The different setting would make the comparison unfair.

Previous vehicle re-ID methods can be mainly summarized into three categories: spatial-attentive feature learning~\cite{orientation,VAMI,RAM,advkeypt,GRF-GGL,DFFMG}, distance metric learning~\cite{GSTE} and embedding learning~\cite{EALN,QD-DLF}.
For spatial-attentive feature learning, proposed methods attempted to guide the network focusing on the regional features which may be useful to distinguish from two vehicles.
RAM~\cite{RAM}, GRF-GLL~\cite{GRF-GGL} and DFFMG~\cite{DFFMG} simply partitioned the images horizontally and vertically into several regions and extract the corresponding regional features; however, when the given images are in different orientations, the features would fail to consistently attends on same parts of vehicle .
To extract orientation-aware features, OIFE~\cite{orientation} and AAVER~\cite{advkeypt} used extra expensive keypoints information to train their orientation-based region proposal network. Yet, they usually lose some informative information like the sticker on the windshield which is not covered by annotated keypoints.
Instead, VAMI~\cite{VAMI} used the viewpoint information to generate representative features of each viewpoint and used them to guide the network producing the viewpoint-aware attention maps and features, but the attention outcomes are not steady.
To sum up, the unfavorable attention masks generated by existing work would hinder the re-ID performance on the benchmarks.
In contrast, our method (\textbf{SPAN w/ CPDM}) achieves clear gains of $\mathbf{7.2}\%$ and $\mathbf{16.7}\%$ for mAP in VeRi-776 and CityFlow-ReID datasets compared to~\cite{GRF-GGL} and~\cite{DFFMG} respectively, indicating that we can benefit from more meaningful attention masks and better utility of global and part features. Also, our method outperforms other state-of-the-arts in both datasets. 

\begin{table}[t]
    \centering
    \mycaption{Comparison with state-of-the-arts re-ID methods on VeRi-776 and CityFlow-ReID dataset($\%$)}{Upper/Lower Group: methods \textbf{without}/\textbf{with} spatial-attentive mechanism. All listed scores are from the methods \textbf{without} adopting spatial-temporal information~\cite{VeRi-776-2} or re-ranking~\cite{re-ranking}.}
    \label{tab:sota}
    \begin{tabular}{l|ccc|ccc}
    \hline
    \multirow{2}{*}{Method} & \multicolumn{3}{c|}{VeRi-776} & \multicolumn{3}{c}{CityFlow-ReID} \\ 
    \cline{2-7}
                                          & R-1   & R-5   & mAP    & R-1  & R-5  & mAP  \\ 
    \hline \hline
    EALN~\cite{EALN}                      & 84.4 & 94.1 & 57.4  & -    & -    & -    \\
    MoV1+BS~\cite{baseline}               & 90.2 & 96.4 & 67.6  & 49.0 & 63.1 & 31.3 \\
    MTML~\cite{MTML}                      & 92.3 & 95.7 & 64.6  & 48.9 & 59.7 & 23.6 \\
    \hline
    OIFE~\cite{orientation}               & 68.3 & 89.7 & 48.0  & -    & -    & -    \\
    VAMI~\cite{VAMI}                      & 77.0 & 90.8 & 50.1  & -    & -    & -    \\
    RAM~\cite{RAM}                        & 88.6 & 94.0 & 61.5  & -    & -    & -    \\
    AAVER~\cite{advkeypt}                 & 89.0 & 94.7 & 61.2  & -    & -    & -    \\
    GRF-GGL~\cite{GRF-GGL}                & 89.4 & 95.0 & 61.7  & -    & -    & -    \\
    DFFMG~\cite{DFFMG}                    & -    & -     & -      & 48.0 & 60.0 & 25.3 \\
    \hline
    \textbf{SPAN w/ CPDM (Ours)} & \textbf{94.0} & \textbf{97.6} & \textbf{68.9} & \textbf{59.5} & \textbf{61.9} & \textbf{42.0}\\
    \hline
    \end{tabular}
\end{table}
\section{Conclusion}
In this paper, we present a novel vehicle re-ID feature learning framework including Semantics-guided Part Attention Network (SPAN) and Co-occurrence Part-attentive Distance Metric (CPDM). 
Our newly-designed SPAN can generate robust and meaningful attention masks on vehicle parts given only the image-level semantic labels for training. This is attributed to the direct supervision by our proposed mask reconstruction loss and two auxiliary losses.
With the help of robust attention masks, the part feature extraction network is able to learn a more discriminative representation. 
Finally, our proposed CPDM can place emphasis on the vehicle parts that co-occurs in two images to better measure the distance between two vehicles.
Both qualitative and quantitative results confirm the quality of generated attention masks and the benefit of dedicated part feature extraction and distance metric.
Experiments also show that our proposed framework performs favorably against existing vehicle re-ID methods.

\section*{Acknowledgment}
This research was supported in part by the Ministry of Science and Technology of Taiwan (MOST 108-2633-E-002-001), National Taiwan University (NTU-108L104039), Intel Corporation, Delta Electronics and Compal Electronics. 

% ---- Bibliography ----
%
% BibTeX users should specify bibliography style 'splncs04'.
% References will then be sorted and formatted in the correct style.
%
\bibliographystyle{splncs04}
\bibliography{egbib}
\end{document}

% \renewcommand\thelinenumber{\color[rgb]{0.2,0.5,0.8}\normalfont\sffamily\scriptsize\arabic{linenumber}\color[rgb]{0,0,0}}
% \renewcommand\makeLineNumber {\hss\thelinenumber\ \hspace{6mm} \rlap{\hskip\textwidth\ \hspace{6.5mm}\thelinenumber}}
% \linenumbers
\pagestyle{headings}
\mainmatter
\def\ECCVSubNumber{3376}  % Insert your submission number here

\title{Orientation-aware Vehicle Re-identification with Semantics-guided Part Attention Network} % Replace with your title

% INITIAL SUBMISSION 
\begin{comment}
\titlerunning{ECCV-20 submission ID \ECCVSubNumber} 
\authorrunning{ECCV-20 submission ID \ECCVSubNumber} 
\author{Anonymous ECCV submission}
\institute{Paper ID \ECCVSubNumber}
\end{comment}
%******************

% CAMERA READY SUBMISSION
%\begin{comment}
\titlerunning{Orientation-aware Vehicle Re-identification with Semantics-guided Network}
% If the paper title is too long for the running head, you can set
% an abbreviated paper title here
%
\author{
Tsai-Shien Chen\inst{1,2} \and
Chih-Ting Liu\inst{1,2} \and
Chih-Wei Wu\inst{1,2} \and
Shao-Yi Chien\inst{1,2}}

\institute{Graduate Institute of Electronic Engineering, National Taiwan University \\ \and
NTU IoX Center, National Taiwan University \\
\email{\{tschen, jackieliu, cwwu\}@media.ee.ntu.edu.tw} \\ 
\email{sychien@ntu.edu.tw}}
%\end{comment}
%******************
\title{Supplementary Material: \\ Orientation-aware Vehicle Re-identification with Semantics-guided Part Attention Network}
\maketitle

\section{Details of Generating Foreground Vehicle Masks}
To get the foreground mask of the whole vehicle, we use a traditional segmentation technique, Grabcut~\cite{grabcut}. However, it requires user to frame the target object out from the whole image for the first stage segmentation, and mark a part of wrong-labeled pixels for obtaining a better result. Yet, neither of them can be done manually owing to the large scale of our dataset. Considering that the input vehicle images in our dataset are all first generated by vehicle detection algorithm, we utilize an automatic method that assumes the pixels on the image border all belong to the background, and therefore we can frame out the object from the border-padding image with the original image size to get the first stage segmentation result. 

To acquire more robust background-removed image, we use the first stage results as target labels to train a segmentation CNN network with one ResNet-50~\cite{resnet} followed by four transposed convolutional layers. But, to avoid the network overfitting on the unstable results generated by Grabcut, after training for a few epochs, we remove the training images with abnormal huge loss, which possibly represent unsatisfactory results done by Grabcut. Finally, we use this trained segmentation CNN network to inference all the data to get the background-removed images.

\begin{figure}[t]
	\centering
    \includegraphics[width=\textwidth]{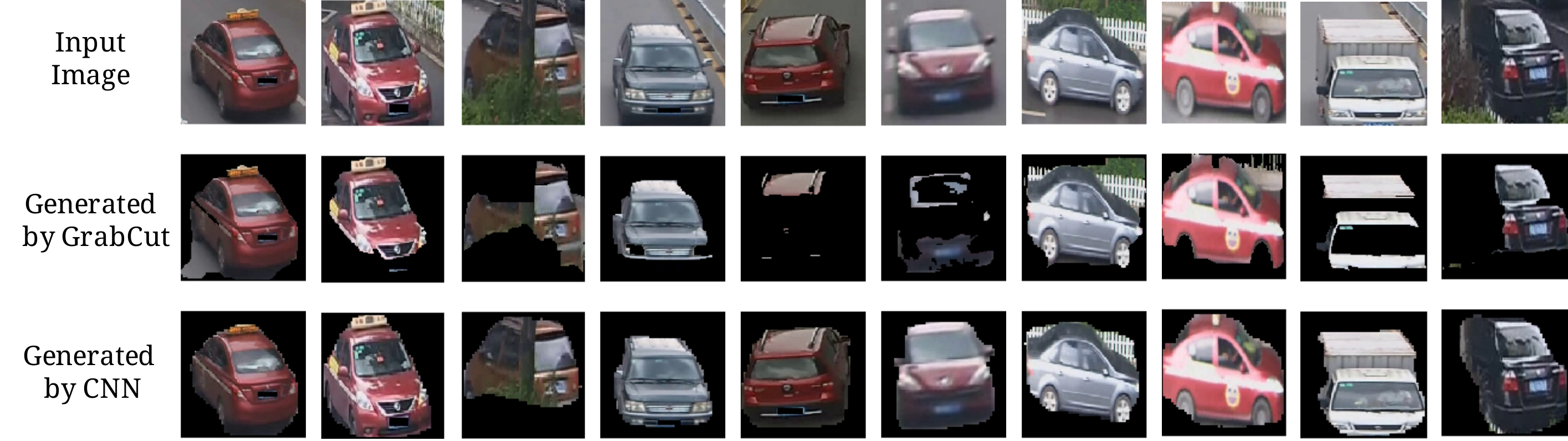}
    \mycaption{Qualitative results of the background-removed images}{The first row shows the input image and the second and third rows are the background-removed images generated by GrabCut and by the segmentation CNN network respectively.}
    \label{fig:bgremove}
\end{figure}

As shown in Fig.~\ref{fig:bgremove}, we visualize some unfavorable background-removed images generated by Grabcut. The first stage results are unsteady; the background region is sometimes mistakenly segmented to foreground while some parts of vehicle which may contain the discriminative features such as wheels and headlamps are sometimes classified to background. In contrast, we can get better results generated by segmentation CNN network.

\newpage
\section{Architecture of our Semantic-guided Part Attention Network}
\begin{figure}[t]
	\centering
    \includegraphics[width=\textwidth]{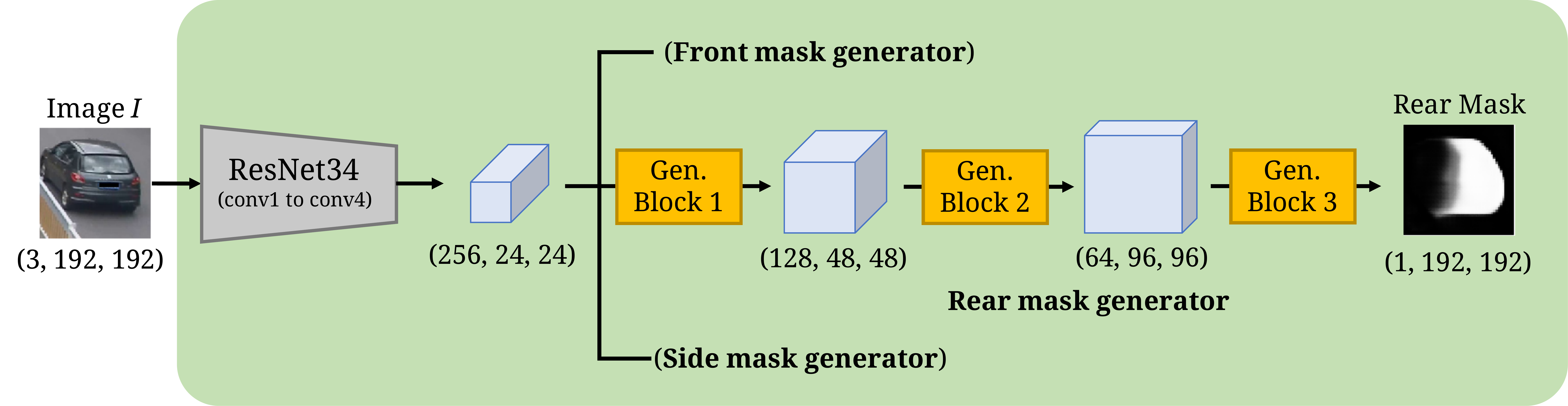}
    \mycaption{Model Architecture of our Semantic-guided Part Attention Network}{}
    \label{fig:SPANstruct}
\end{figure}

The network architecture of our proposed Semantic-guided Part Attention Network (SPAN) is shown in Fig.~\ref{fig:SPANstruct}. It consists of a feature extractor which is the $conv1$ to $conv4$ in ResNet-34~\cite{resnet} ($CNN_{mask}$ in the main paper) and three mask generators (front, rear and side) with the same architecture, which only the rear mask generator is illustrated in details. Each generator contains three generative blocks (Gen.\ Block) and each block includes one transposed convolutional layer, batchnorm and ReLU layer. Considering that too powerful CNN model and extensive receptive field would lead to unexpected training results as described in the main paper, we only use the former four blocks in ResNet-34.

\section{Selection of Hyper-parameters in Loss Functions.}
\begin{figure}[t]
	\centering
    \includegraphics[width=\textwidth]{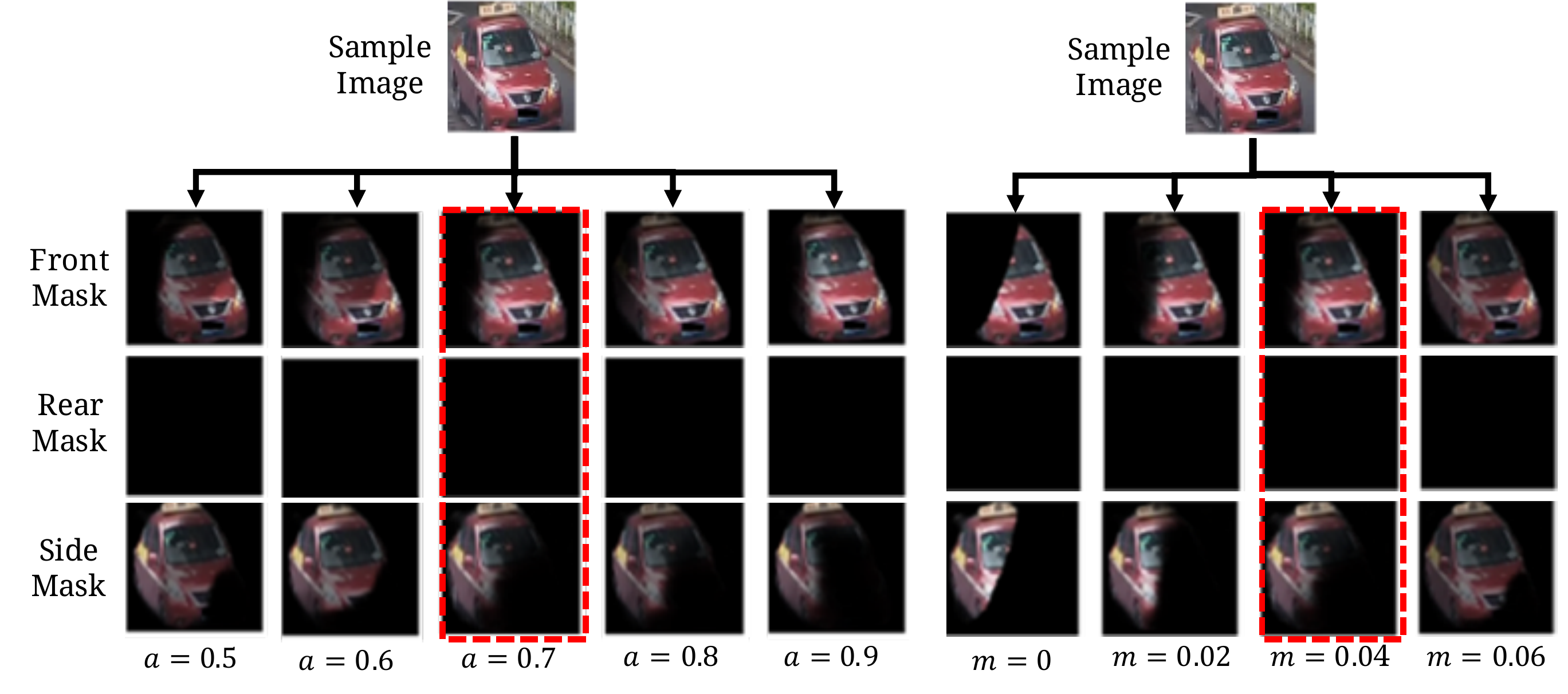}
    \mycaption{Analysis of the hyper-parameters: max area ratio $a$ in $\mathcal{L}_{area}$ and margin $m$ in $\mathcal{L}_{div}$}{We block the final selection of parameters in the red frames.}
    \label{fig:parameters}
\end{figure}

We adopt three loss functions ($\mathcal{L}_{recon}$, $\mathcal{L}_{area}$ and $\mathcal{L}_{div}$) to supervise the training of our SPAN model. When computing the losses, there are two hyper-parameters should be selected, including max area ratio $a$ in $\mathcal{L}_{area}$ and margin $m$ in $\mathcal{L}_{div}$. The physical meaning and selection have been discussed in the main paper.
To select the hyper-parameters, we split a validation set out from the original training set of VeRi-776 dataset~\cite{VeRi-776-1,VeRi-776-2} and observe the quality of generated attention masks of sampled images from validation set. 
We adjust one of the hyper-parameter while the other is fixed. The experiment results are shown in Fig~\ref{fig:parameters}.

The ideal part attention masks should cover all regional features which are belonging to their views while exclude the others. Take the results in Fig~\ref{fig:parameters} as example, the front masks of $a = 0.8$ and $m = 0.06$ mistakenly include side views and the front mask of $m = 0.02$ incorrectly loses part of front view. Hence, based on the experiment results, we finally choose $a = 0.7$ for the visible views of two-view images and $m = 0.04$ for two adjacent views. The complete selection of hyper-parameters is shown in Table~\ref{tab:para_area} and~\ref{tab:para_div}.

\begin{table}[t!]
    \begin{minipage}{.5\linewidth}
        \centering
        \mycaption{Parameters of $\mathcal{L}_{area}$}{}
        \begin{tabular}{ l|ccc }
         \hline
         \multirow{2}{*}{viewpoint} &
         \multicolumn{3}{c}{max area ratio $a_l$} \\\cline{2-4}
         \multicolumn{1}{c|}{} & front & rear & side \\
         \hline \hline
         front       & 1   & 0   & 0 \\ 
         rear        & 0   & 1   & 0 \\
         side        & 0   & 0   & 1 \\
         front-side  & 0.7 & 0   & 0.7 \\
         rear-side   & 0   & 0.7 & 0.7 \\
         \hline
        \end{tabular}
        \label{tab:para_area}
    \end{minipage}
    \begin{minipage}{.5\linewidth}
        \centering
        \mycaption{Parameters of $\mathcal{L}_{div}$}{}
        \begin{tabular}{ l|c } 
         \hline
         view pair & margin $m$\\
         \hline \hline
         front, rear & 0    \\ 
         front, side & 0.04 \\
         rear, side & 0.04 \\
         \hline
        \end{tabular}
       \label{tab:para_div}
    \end{minipage}
\end{table}

\section{Other Qualitative Results of the Generated Part Masks}
\begin{figure}[t]
	\centering
    \includegraphics[width=\textwidth]{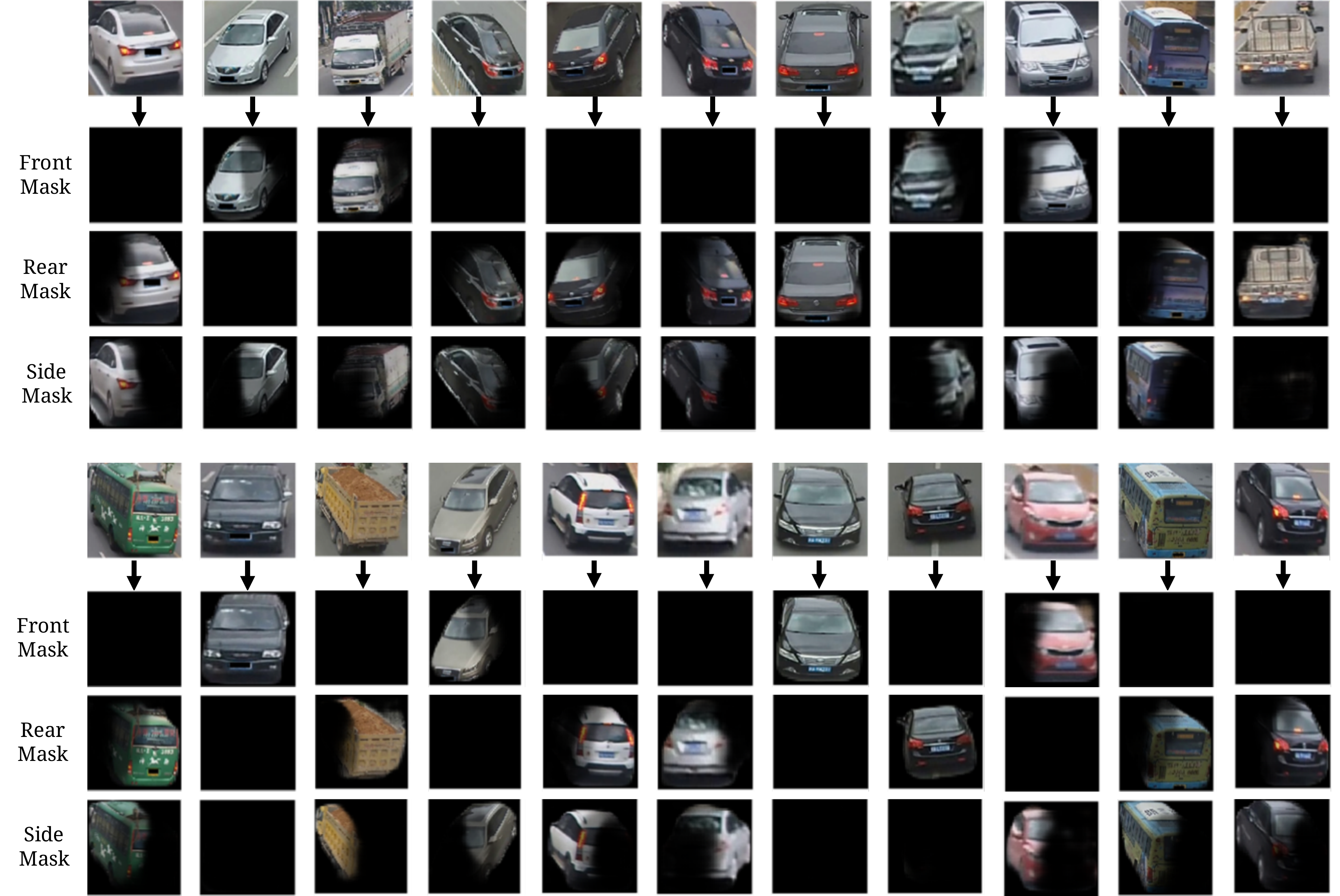}
    \mycaption{Qualitative Results of Generated Part Masks}{}
    \label{fig:qualitative}
\end{figure}

To verify the robustness of SPAN, we show more qualitative results of the generated part masks in Fig.~\ref{fig:qualitative}. It is worth mentioning that the input images are all randomly chosen from the whole VeRi-776 dataset~\cite{VeRi-776-1,VeRi-776-2} without manually selected.

% ---- Bibliography ----
%
% BibTeX users should specify bibliography style 'splncs04'.
% References will then be sorted and formatted in the correct style.
%
\newpage
\bibliographystyle{splncs04}
\bibliography{egbib}